\documentclass{article}
\usepackage{spconf,amsmath,graphicx}
\usepackage{algorithmic}
\usepackage{algorithm}
\usepackage{array}
\usepackage{afterpage}
\usepackage[caption=false,font=normalsize,labelfont=sf,textfont=sf]{subfig}
\usepackage{textcomp}
\usepackage{stfloats}
\usepackage{url}
\usepackage{verbatim}
\usepackage{graphicx}
\usepackage{cite}
\usepackage{bbm}
\usepackage{amssymb,amsmath}
\usepackage{xcolor}
\usepackage{booktabs}
\usepackage{url}
\usepackage{lscape}

\newcommand\an[1]{\textcolor{magenta}{[Aviv: #1]}}

\newcommand{\bfy}{\mathbf{y}}
\newcommand{\bfc}{\mathbf{c}}
\newcommand{\bfx}{\mathbf{x}}
\newcommand{\rmp}{\mathrm{P}}

\newcommand{\eqeq}{\mathrel{=}\mathrel{=}}

\definecolor{general}{HTML}{006795}  
\definecolor{jargon_imp}{HTML}{008000} 
\definecolor{jargon}{HTML}{EE4B2B}
\definecolor{emerald}{HTML}{00A99D} 

\title{COMBINING LANGUAGE MODELS FOR SPECIALIZED DOMAINS: A COLORFUL APPROACH}

\name{D. Eitan, M. Pirchi, N. Glazer, S. Meital, G. Ayach, G. Krendel, A. Shamsian, A. Navon, G. Hetz, J. Keshet}
\address{aiOla Research}
\begin{document}
%
\maketitle
\begin{abstract}
General purpose 
language models (LMs) encounter difficulties when processing domain-specific jargon and terminology, which are frequently utilized in specialized fields such as medicine or industrial settings. Moreover, they often find it challenging to interpret mixed speech that blends general language with specialized jargon. This poses a challenge for automatic speech recognition systems operating within these specific domains. In this work, we introduce a novel approach that integrates domain-specific or secondary LM into general-purpose LM. This strategy involves labeling, or ``coloring'', each word to indicate its association with either the general or the domain-specific LM. We develop an optimized algorithm that enhances the beam search algorithm to effectively handle inferences involving colored words. Our evaluations indicate that this approach is highly effective in integrating jargon into language tasks. Notably, our method substantially lowers the error rate for domain-specific words without compromising performance in the general domain.
\end{abstract}
\section{Introduction}
\label{sec:intro}
Specialized jargon is crucial in numerous tasks and real-world scenarios, facilitating precise and efficient communication among experts or professionals. Jargon language is prevalent in sectors such as law, industry, business, and healthcare. The sentence ``The patient suffered from \emph{formication} due to  \emph{neuropathy} caused by \emph{diabetes mellitus}'' is an example from healthcare jargon language where the jargon term are highlighted in italics. In the subsequent sections, we will use the term \emph{mixed speech} to refer to spoken expressions that combine general language with specialized jargon.

Jargon language poses significant challenges for automatic speech recognition (ASR) and natural language processing (NLP) systems, which are predominantly trained on more general language datasets. While some jargon words and terms are publicly accessible  (e.g., flight control or some medical terms) and hence can be used in training or fine-tuning LMs, many jargon words are used exclusively within industrial companies. As a result, they are not available for training in the widely used general-purpose language models.

\begin{figure*}
  \centering
  \includegraphics[width=\textwidth, height=4.5cm]{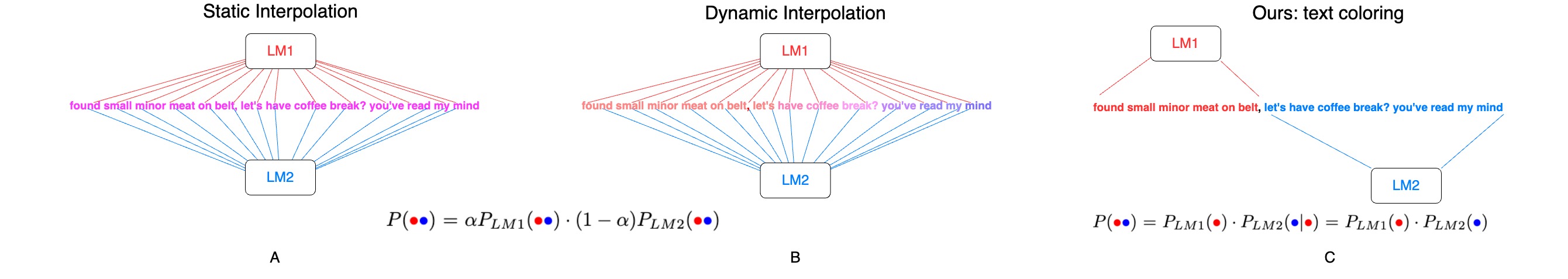} 
  \caption{Illustration of the probability calculation in language models for mixed speech using different approaches: (A) static interpolation, and (B) dynamic interpolation — both methods facilitate interpolation between LMs. Conversely, in approach (C), which is our proposed method, each LM is responsible for handling either the general or the jargon segment exclusively.}
  \label{fig:interpolation_ours_bayes}
\end{figure*}


One common approach to address this challenge involves interpolating between multiple LMs, each specializing in a different domain and estimated using distinct corpus \cite{klakow1998log, broman2005methods, lv2009novel, sak2013mixture}. However, this approach suffers from three limitations. First, it often relies on fixed interpolation coefficients, which may be sub-optimal~\cite{hsu2007generalized,
pusateri2019connecting, loglinearInterpolation_NN}. Second, it does not address mixed speech within a sentence, resulting in low probability scores for such sentences from all LMs. Third, utilizing the average outputs of several LMs can lead to an over-smoothing effect \cite{broman2005methods}. 

The problem of incorporating several LMs also relates to topic-model LMs \cite{gildea1999topic, hsu2006style,  watanabe2011topic}. Topic-model LM is a mixture model where each mixture component is represented by a topic-dependent word probability, and each mixture weight corresponds to a topic proportion probability. These methods assume an underlying parametric distribution and are often implemented using a unigram word probabilities \cite{watanabe2011topic}. 

In this study, we introduce a novel method for
integrating multiple LMs by dynamically switching between general and jargon-specific models, mitigating the limitations of
previous interpolation methods. 
We demonstrate the effectiveness of our approach in the realm of ASR, 
which has seen remarkable advancements in recent years. This development can be attributed to the emergence of self-supervised representation techniques \cite{baevski2020wav2vec, hsu2021hubert}.
However, 
precise language modeling is crucial for achieving high accuracy rates in ASR systems. 

We describe an efficient procedure for implementing our method, based on the beam search algorithm. As our approach requires only minor changes to the original algorithm, it is easy to integrate into existing systems. Our approach provides a practical solution for handling domain-specific jargon in speech recognition, with important implications for a wide range of ASR applications that rely on the accurate transcription of specialized terminology. 
Importantly, our approach can easily extend and be utilized in other NLP tasks, such as MT \cite{leblond2021machine, weller2022end} and text generation \cite{brown2020language}.


This paper makes the following contributions: (i) it introduces a novel approach for incorporating general and domain-specific LMs; (ii) it develops an efficient and scalable algorithm, grounded in beam search techniques, for approximating the proposed method; and (iii) it undertakes a set of experiments to empirically validate the efficacy of our approach.

\section{Method}
\label{sec:method}

A language model is defined as the probability of a sequence of words. Denote the sequence of words by $\bfy=(y_1, \ldots, y_T)$. Each word $y_t$ is associated to one or more of the $C$ vocabularies $\mathcal{V}^1, \ldots, \mathcal{V}^C$. 
For simplicity of presentation we will discus our solution for combining two LMs namely a general and jargon  models.   
We denote the input sequence by $\mathbf{x}=(x_1,\ldots,x_L)$. This sequence can be a speech utterance for ASR ~\cite{baevski2020wav2vec, hsu2021hubert}, a sequence of word tokens for (conditional) language generation~\cite{brown2020language}, or 
a sequence of words from a different language in the case of MT~\cite{leblond2021machine}. Let $\mathcal{Y}(\mathbf{x})$ denote the set of all valid output sequences for input $\mathbf{x}$. Given an input $\mathbf{x}$, our goal is to obta  in the maximum a posteriori (MAP),
\begin{align}\label{eq:map}
\mathbf{\hat{y}}=\arg\max_{\mathbf{y}\in\mathcal{Y}(\mathbf{x})} \mathrm{P}(\mathbf{y}\mid \mathbf{x})
\end{align}

In this work, we focus on additive scoring rules as commonly used in MAP decoding of neural sequence-to-sequence models. Concretely, at each position $t$ we wish to obtain,
\begin{equation}\label{eq:dec_prob}
\hat{y}_t=\arg\max_{y_t} \mathrm{P}(y_t\mid \mathbf{y}_{<t},\mathbf{x})
\end{equation}
where $\bfy_{<t}=(y_1,\ldots,y_{t-1})$. 

\medskip
We introduce an additional sequence $\bfc=(c_1,...,c_T)$ where $c_t \in [C]$, that controls the source from which $y_t$ is derived, i.e., $c_t=j$ if $y_t \in \mathcal{V}^j$. We refer to $\bfc$ as the coloring variable. Our approach modifies Eq.~\eqref{eq:map} to account for the additional coloring variable and jointly optimize $\bfy,\bfc$. Assuming $c_t$ is independent of the colored history $(\bfy,\bfc)_{<t} = ((y_1, c_1), \ldots, (y_{t-1}, c_{t-1}))$ we write,
\begin{equation}
\begin{aligned}\label{eq:decoding}
(\hat{\mathbf{y}},\hat{\mathbf{c}})&=\arg\max_{\mathbf{y},\mathbf{c}} \prod_t \mathrm{P}(y_t,c_t\mid (\mathbf{y},\mathbf{c})_{<t},\mathbf{x})\\
    &=\arg\max_{\mathbf{y},\mathbf{c}}\prod_t \mathrm{P}(c_t\mid (\mathbf{y},\mathbf{c})_{<t}) \mathrm{P}(y_t\mid \mathbf{c}_{\leq t},\mathbf{y}_{<t},\mathbf{x})\\
    &=\arg\max_{\mathbf{y},\mathbf{c}}\prod_t \mathrm{P}(c_t) \mathrm{P}(y_t\mid \mathbf{c}_{\leq t},\mathbf{y}_{<t},\mathbf{x})~,
\end{aligned}
\end{equation}
 where the last equality follows from our independence assumption. For each time $t$, our method optimizes for the upcoming word and its source lexicon, conditioned on the ``colored'' history. In the experiments, we fixed coloring probabilities $\rmp(c_t=j)=\rmp(c=j)=1/C$ for each $j\in[C]$. We choose to use a uniform distribution for the colors as a reasonable assumption, given the scarcity of mixed speech texts. 

In order to understand the advantages of our approach, let's examine the challenge of integrating jargon LMs into a more general LM. Assuming the general LM is optimized for standard language, and the jargon model is optimized for domain-specific language, any interpolation method will fall short when it comes to mixed speech. However, our approach dynamically switches between the two LMs, allowing for an accurate estimation of the likelihood of mixed sentences. Additionally, our method assigns the source lexicon for each word in the decoded text. This supplementary information can be advantageous for downstream NLP tasks.

\begin{algorithm}[t]
    \caption{CTC Beam Search with Coloring}\small\label{alg:beam}
    \begin{algorithmic}[H]
    \STATE {\bfseries Input:} AM output scores matrix $S$, beam width $w$.
    \STATE {\bfseries Initialize:} $B\gets \{\varnothing \}$, $\rmp^{-}(\varnothing, 0)=1$     
    \FOR{$t=1,...,L$}
        \STATE $\hat{B}\gets GetBestBeams(B, w)$
        \STATE $B\gets \{\}$
        \FOR{$\tilde{\bfy}=(\bfy,\bfc) \in \hat{B}$}
            \IF{$\bfy \neq \varnothing$}
                \STATE $\rmp^{+}(\bfy,t) \gets  \rmp^{+}(\bfy,t-1)\cdot S[y_{-1}, t]$
            \ENDIF
            \STATE $\rmp^{-}(\bfy,t) \gets \rmp^{\text{tot}}(\bfy,t-1) \cdot S[-, t]$
            \STATE $B\gets B\cup \{\tilde{\bfy}\}$
            \STATE \textcolor{jargon_imp}{$NextChars \gets GetNextChars(\tilde{\bfy})$}
            \FOR{$(\ell, c) \in NextChars$}
            \STATE $\bfy' \gets  \bfy \circ \ell$
            \STATE \textcolor{jargon_imp}{$\bfc' \gets  \bfc \circ c$}
            \STATE \textcolor{jargon_imp}{$\tilde{\bfy}' \gets  (\bfy', \bfc')$}
            \STATE \textcolor{jargon_imp}{$\rmp^{\text{text}}(\tilde{\bfy}')\gets ScoreBeam(\tilde{\bfy},\ell,c)$}
            \IF{$y_t \eqeq \ell$}
                \STATE $\rmp^{+} (\bfy',t) \gets  \rmp^{-}(\bfy,t-1) \cdot S[i_\ell ,t]$
            \ELSE
                \STATE $\rmp^{+} (\bfy',t) \gets  \rmp^{\text{tot}}(\bfy,t-1) \cdot S[i_\ell ,t]$
            \ENDIF
            \STATE $B\gets B\cup \{\tilde{\bfy}'\}$
            \ENDFOR
        \ENDFOR
    \ENDFOR
    \STATE {\bfseries Return:} $GetBestBeams(B, 1)$
\end{algorithmic}
\end{algorithm}

\section{Implementation Details}\label{sec:imp_details}
In this section, we describe an approach to utilize the beam search algorithm and efficiently scaling with the number of lexicons, denoted by $C$.

It may not be possible to solve the exact exhaustive search of the decoding in Eq.~\eqref{eq:decoding}. Therefore, beam-search decoding approximation is often utilized~\cite{jain2019rnn, yang2019xlnet}. 

In our settings, the input for the beam search process is the acoustic model's output, i.e., the probability scores (logits) for each token (characters in our case) at each time step. To accommodate the different lexicon sources, we identify each token score with the corresponding tokens in all lexicons. 
During the beam search, each word's first letter can come from either of the lexicons, once the first letter has been set, only letters from the same lexicon can be appended, thus alleviating the computational cost.

We now provide a detailed description of the implementation details and modifications to the beam search algorithm. For simplicity and ease of exposition, we consider having two lexicons, a general and a jargon-specific, denoted $\mathcal{V}^G, \mathcal{V}^J$, respectively. Our method, however, trivially extends to multiple lexicons. We denote the characters set by $\mathcal{A}$ of size $|\mathcal{A}|=K$. To achieve word coloring and identify jargon terms, we modify the character set of the jargon lexicon. We denote the jargon characters set  $\mathcal{A}'$, and identify each $\ell'\in \mathcal{A}'$ with $\ell\in \mathcal{A}$ in a one-to-one manner by color each character in $\mathcal{A}$. We further identify each character $\ell\in\mathcal{A}$ with an integer $i_{\ell}\in \{1, 2, \ldots, K\}$ such that $i_\ell=i_{\ell'}$. The importance of having $\mathcal{A}$ and $\mathcal{A}'$ as disjoint sets lies in the fact that the n-grams of the models are also necessarily disjoint in this scenario. This makes it straightforward to merge the original n-gram models simply by appending the n-grams. We let $\varnothing$ denote the empty sequence, and $S=S(\bfx)\in \mathbb{R}^{L\times (K+1)}$ the output from the acoustic model ($K$ characters and one blank symbol denoted $-$). $S$ is arranged such that the $i_{\ell}$th column corresponds to $\ell\in\mathcal{A}$.
Furthermore, let $\rmp^{-}$, $\rmp^{+}$ and $\rmp^{\text{tot}}$ denote the \emph{blank}, \emph{non-blank} and \emph{total} probabilities, with $\rmp^{\text{tot}}=\rmp^{-}+\rmp^{+}$. We denote  $\tilde{\bfy}$ the joint sequence $(\bfy,\bfc)$, and let $\circ$ denote the concatenation operation. Last, we denote by $y_{-1}$ the last character of the sequence $\bfy$. 

We define three functions $GetNextChars$, $ScoreBeam$ and $GetBestBeams$. The function $GetNextChars(\tilde{\bfy})$ return a set of possible characters given $\tilde{\bfy}=(\bfy,\bfc)$. At each time $t$, it returns the characters corresponding to lexicon $c_{t-1}$ if the current beam ends with a sub-word. Otherwise, it returns all possible characters. The function $ScoreBeam(\tilde{\bfy},\ell,c)$ returns the probability of seeing $(\ell,c)$ as an extension of the beam $\tilde{\bfy}$. Using $ScoreBeam$, we define $\rmp^{\text{text}}(\tilde{\bfy}')=ScoreBeam(\tilde{\bfy},\ell,c)$ for $\tilde{\bfy}'=(\bfy\circ \ell, \bfc \circ c)$. 
Finally, the function $GetBestBeams(B,M)$ returns the top $M$ beams from $B$ according to $\rmp^{\text{tot}}\cdot \rmp^{\text{text}}$.

Alg.~\ref{alg:beam} describes our modification to the standard CTC beam search algorithm. The modifications are marked \emph{\textcolor{jargon_imp}{green}}. Our method involves making slight adjustments to the original CTC beam search algorithm. Furthermore, the number of beams in our approach should remain the same as the original CTC beam search, as $|GetNextChars(\tilde{y})| = K$ for each sub-word. As a result, our approach can be easily scaled with the number of lexicons $C$.


\begin{table*}[t]
\centering
\small
\caption{Incorporating jargon LMs using four datasets: Industrial English, two medical datasets~\cite{fareez2022dataset,mts-dialog}, and Industrial Thai.}
\vspace{0.2cm}
\begin{tabular}{lcccccccc}
\toprule
 &  \multicolumn{2}{c}{Industrial English} & \multicolumn{2}{c}{Medical~\cite{fareez2022dataset}} & \multicolumn{2}{c}{Medical~\cite{mts-dialog}} &  \multicolumn{2}{c}{Industrial Thai} \\
\cmidrule(lr){2-3} \cmidrule(lr){4-5} \cmidrule(lr){6-7} \cmidrule(lr){8-9}
 &   CER $\downarrow$ &  WER $\downarrow$  &   CER $\downarrow$ &  WER $\downarrow$ &   CER $\downarrow$ &  WER $\downarrow$ &  CER $\downarrow$ &  WER $\downarrow$ \\
\midrule
General LM & $12.6$  & $36.8$ &  $6.1$ &  $13.0$ & $12.1$ & $23.6$ & $6.0$  & $10.6$  \\
Jargon LM & $9.6$  & $21.9$ &  $24.9$ &  $62.7$ & $21.5$ & $56.4$ & $22.1$  & $51.8$ \\
Linear interpolation & $6.9$  & $15.9$ &  $6.2$ &  $12.8$ & $6.2$ & $12.6$ & $5.7$  & $\mathbf{9.9}$ \\
Log-linear interpolation & $10.2$  & $23.9$ &  $6.1$ &  $13.7$ & $7.2$ & $17.3$ & $6.1$  & $11.3$ \\
Bin estimation~\cite{broman2005methods} & $9.2$  & $22.6$ &  $8.2$ &  $21.5$ & $10.9$ & $39.8$ & $6.2$  & $12.6$ \\
Bayes~\cite{bayesinterp2011} & $9.0$  & $23.3$ &  $5.6$ &  $12.8$ & $7.4$ & $18.3$ & $6.0$  & $11.5$ \\
\midrule
Ours & $\mathbf{5.7}$  & $\mathbf{11.9}$ &   $\mathbf{6.1}$ &  $\mathbf{12.3}$ &   $\mathbf{5.8}$ &  $\mathbf{11.4}$ & $\mathbf{5.5}$  & $\mathbf{9.9}$ \\
\bottomrule
\end{tabular}
\vspace{-0.2cm}
\label{tab:jargon}
\end{table*}

\section{Experiments}\label{sec:exp}

In this section, we compare our method with natural baselines for combining language models. 
In all experiments, we used two lexicons, a general and a domain-specific or secondary LM. 
We denote the probability density functions under the two LMs as $\rmp_G$ and $\rmp_J$.


\noindent\textbf{Datasets.} We assess our approach using four datasets: (i) \emph{Industrial English}: $\sim\!\!\!2.5$ hours, $267$ audio files, featuring Australian English utterances on mechanical equipment conditions (jargon) and inspector discussions (general) with machinery background noise. (ii) \emph{Industrial Thai}: $\sim\!\!\!1$ hour, $90$ Thai audio files, similar to the Industrial English dataset. (iii) \emph{Medical} \cite{fareez2022dataset}: $\sim\!\!55$ hours of simulated patient-physician interviews. We segmented recordings into 30-second audio files, using $\sim\!\!1.5$ hours for our experiment (0.5 hours for validation, 1 hour for testing). (iv) \emph{Medical} \cite{mts-dialog}: $1,700$ brief doctor-patient conversations. We extracted 400 sentences for medical terms, created a 3-gram LM, and augmented the dataset with 401 audio files (totaling $80$ minutes). For evaluation, we split it into a validation set (149 files, $\sim\!\!25$ minutes) and a test set (252 files, $\sim\!\!55$ minutes).

\noindent\textbf{Comparison Methods.} We evaluate several baseline approaches for combining language models:
(i) \emph{Linear Interpolation}: Combines language models using $\lambda \mathrm{P}_J+(1-\lambda)\mathrm{P}_G$ for $\lambda \in [0,1]$.
(ii) \emph{Log-Linear Interpolation}~\cite{klakow1998log}: Interpolates log probabilities: $ \mathrm{P}_J^{\lambda}\mathrm{P}_G^{(1-\lambda)}$.
(iii) \emph{Bin Estimation Method}~\cite{broman2005methods}: Maps word probabilities from different LMs into a single probability by binning the space and calibrating the output.
(iv) \emph{Bayes Approach}~\cite{bayesinterp2011}: A dynamic interpolation method based on Bayes' theorem, assigning interpolation weights for each word.
(v) \emph{General LM}: Uses only the general LM.
(vi) \emph{Jargon LM}: Uses only the jargon LM.
(vii) \emph{Ours}: Our proposed approach as described in Section~\ref{sec:method}.



\begin{table*}[t!]
\centering
    \caption{Qualitative examples 
    from the medical dataset~\cite{mts-dialog}. 
    General LM marked in \textcolor{general}{\emph{blue}}, while domain-specific LM in \textcolor{jargon}{\textbf{\emph{red}}}.}
    \vskip 0.11in
\scriptsize
\begin{tabular}{{lp{4.6cm}p{4.6cm}p{4.6cm}}}
\toprule
 &  \multicolumn{1}{c}{Ground Truth} & \multicolumn{1}{c}{Linear Interpolation} & \multicolumn{1}{c}{Ours} \\
\midrule
 Success & She  had no fever or chills \textbf{cough} congestion nausea vomiting chest pain chest pressure & She  had no fever or chills of congestion nausea vomiting chest pain chest pressure & 
 \textcolor{general}{She  had no} \textcolor{jargon}{\textbf{fever}} \textcolor{general}{or} \textcolor{jargon}{\textbf{chills cough congestion nausea vomiting chest pain chest pressure}} \\
 \\
 Success & He has had years in which he did better on \textbf{clozaril} and also his other medications & He has had year in which he did better on care and other medications & \textcolor{general}{He has had years in which he did better on} \textcolor{jargon}{\textbf{clozaril}} \textcolor{general}{and also his other} \textcolor{jargon}{\textbf{medications}}  \\
 \\
 Failure & He has not been treated for diabetes since his last  weight since he stopped taking \textbf{zyprexa} & He has not been treated for diabetes since his  weight since he stopped taking the pre &  \textcolor{general}{He has not been treated for} \textcolor{jargon}{\textbf{diabetes}} \textcolor{general}{since his last weight since he stopped taking the prex}  \\
\bottomrule
\end{tabular}
\vspace{-0.2cm}
\label{tab:qualitative}
\end{table*}

\noindent\textbf{Hyperparameter Search (HP).} We performed grid searches with validation splits to optimize hyperparameters. For all methods, we explored $\alpha$ and $\beta$ in $\{0.5, 0.75, 1.0, 1.25, 1.5\}$, representing language model contribution and text length scaling. As well as unknown word penalties in $\{-10, -50\}$ for each LM.
For linear and log-linear interpolation, we tuned $\lambda\in\{0.25, 0.5, 0.75\}$. 
For the bin estimation method~\cite{broman2005methods}, we considered the number of bins in $\{53, 100\}$ (where $53$ was optimal~\cite{broman2005methods}).
In our method, we explored an unknown sub-word penalty in $\{-7, -5, -3, -1, 0\}$.


\vspace{-0.2cm}
\subsection{Incorporation Domain-Specific Jargon}\label{sec:inc_jargon}

Here we evaluate our method on the task of incorporating domain-specific jargon into general LMs. We used the (i) Industrial EN, the (ii) medical dataset~\cite{fareez2022dataset} and (iii) medical dataset~\cite{mts-dialog}. We employed a pre-trained XLSR-Wav2Vec  2.0~\cite{conneau2020unsupervised} based model fine-tuned using the Common Voice dataset~\cite{commonvoice:2020}\footnote{ \url{https://huggingface.co/jonatasgrosman/wav2vec2-large-xlsr-} \url{53-english}.}. We use a $3-$gram English LM\footnote{ \url{https://catalog.ngc.nvidia.com/orgs/nvidia/teams/tlt-jarvis/models} \url{/speechtotext_english_lm}.} 
as the general LM. Furthermore, we constructed domain-specific $3-$gram LMs using held-out data. (iv) Industrial TH dataset containing Thai sentences with specialized terms. We used a pre-trained Thai XLSR-Wav2Vec 2.0 based model~\cite{phatthiyaphaibun2022thai} fine-tuned using the Thai Common Voice corpus V8~\cite{commonvoice:2020}. 
We trained two $3$-gram LMs; The first was trained on a standard Thai, while the second  contained factory terms.

The results are presented in Table~\ref{tab:jargon}. Our method achieves a significant reduction in the WER and CER compared with the natural baselines for combining language models. Notably, the significant error reduction in the Industrial dataset which was recorded in a challenging acoustic environment.

\vspace{-0.2cm}
\subsection{Qualitative Examples}\label{sec:qualitative}


To gain insights into the successes and failures of our method, this section presents a collection of qualitative examples that showcase the efficacy of our proposed approach. The examples, presented in Table~\ref{tab:qualitative}, are selected from the medical dataset proposed in~\cite{mts-dialog}. 
The first two examples demonstrate how our method accurately produces the ground truth (GT) transcription. On the other hand, the linear interpolation model fails to produce some of the domain-specific terms; We highlight these terms in the GT column for clarity. 
The last example presents a failure case: both our approach and the linear interpolation method fail to produce the domain-specific word \emph{zyprexa}.

\vspace{-0.4cm}
\section{Conclusions}
\vspace{-0.2cm}
In this work, we introduce a novel method for combining language models, demonstrating its effectiveness in ASR with domain-specific jargon. We present an efficient algorithm based on the common beam search, requiring only minor modifications to integrate into ASR systems. Our approach offers a practical solution for accurate transcription of specialized terminology in diverse ASR applications.

\bibliographystyle{IEEEtran}
\bibliography{jargon_decoder}

\vspace{11pt}

\vfill

\end{document}